\documentclass{article}

\usepackage[preprint]{neurips_2024}
\usepackage[utf8]{inputenc} 
\usepackage[T1]{fontenc}    
\usepackage{hyperref}       
\usepackage{url}            
\usepackage{booktabs}       
\usepackage{amsfonts}       
\usepackage{nicefrac}       
\usepackage{microtype}      
\usepackage{xcolor}         

\usepackage{multirow}
\usepackage{caption}
\usepackage{subcaption}
\usepackage{graphicx}
\usepackage{placeins}

\usepackage{natbib}
\setcitestyle{numbers}

\title{GSAVS: Gaussian Splatting-based Autonomous Vehicle Simulator}

\author{
  Rami Wilson\\
  Computer Science and Engineering\\
  UC Santa Cruz\\
  \texttt{ramewils@ucsc.edu} \\
}

\begin{document}

\maketitle

\begin{abstract}
Modern autonomous vehicle simulators feature an ever-growing library of assets, including vehicles, buildings, roads, pedestrians, and more. While this level of customization proves beneficial when creating virtual urban environments, this process becomes cumbersome when intending to train within a digital twin or a duplicate of a real scene. Gaussian splatting emerged as a powerful technique in scene reconstruction and novel view synthesis, boasting high fidelity and rendering speeds. In this paper, we introduce GSAVS, an autonomous vehicle simulator that supports the creation and development of autonomous vehicle models. Every asset within the simulator is a 3D Gaussian splat, including the vehicles and the environment. However, the simulator runs within a classical 3D engine, rendering 3D Gaussian splats in real-time. This allows the simulator to utilize the photorealism that 3D Gaussian splatting boasts while providing the customization and ease of use of a classical 3D engine.
\end{abstract}

\section{Introduction}
Simulating dangerous scenes is necessary to create robust autonomous driving models. Since some driving scenarios are too dangerous to recreate in reality (car crashes, close calls, etc.), a simulation is a safe alternative to provide enough training data variety for autonomous vehicle models.\cite{NIPS1988_812b4ba2} This further improves the robustness of autonomous vehicle models and allows them to navigate a greater variety of driving scenarios.

Current state-of-the-art simulators, such as CARLA \cite{conf/corl/DosovitskiyRCLK17}, excel at representing the complexity of urban environments. Classic mesh-based approaches are utilized to represent scenes, with separate assets to represent every component. This allows for the customization of assets to fit any use case, including scene parameters such as lighting, traffic rules, intersections, and any other parameters that can potentially affect the driving model.

The use of standard 3D file formats is what gives these simulators their extensibility. For example, in CARLA, while users can use the generated assets and scenes, users also have the ability to create and import new vehicles and maps in the standard .fbx format. \cite{conf/corl/DosovitskiyRCLK17} This allows for the creation of custom maps and scenes with the corresponding assets, making simulators like CARLA a platform capable of representing scenes with a high level of granular control.

The goal of any autonomous vehicle simulator is to train a model that is capable of bridging the sim-to-real gap. That is, a model's performance in a virtual simulator should ideally transfer to the real world. However, in practice, we find that is not the case. For example, the ego vehicle trained in a simulator may have different dynamics than the real-world vehicle. These dynamics can include different lighting, different agents, noise in the input data, etc. This means that learned actions are usually not executed in the real world as the model expects, and even states that the model perceives could be different from training \cite{morris2024slugmobiletestbenchrl}.

In order to balance performance, simulators compensate by rendering assets at a lower fidelity. Especially in the context of real-time rendering of simulations, balancing photorealism with performance is computationally expensive. Additionally, training exclusively with virtual data captured within the simulator has proved to be insufficient in training a model that is robust enough to transfer well to the real world. \cite{VOOGD20231510} This also requires a lot of fine-tuning of the model to the real world, which can be time-consuming and costly. \cite{10242366} To combat this, high-fidelity digital twins have proven to be a viable solution. This allows the model to train in an environment that is closer to the real world, further making the model more robust and progressing towards closing the sim-to-real gap.

In this paper, we introduce GSAVS, a Constrained Gaussian Splatting-based autonomous vehicle simulator. GSAVS has been built to simplify the asset creation process while maintaining high rendering speeds and fidelity. This is because every asset, including the environment and vehicles, is rendered by the simulator as a 3D Gaussian splat in real time. However, the simulator runs entirely within the Unity \cite{unity} engine, wrapping around the 3D Gaussian splat components. Using a classic 3D engine allows for the customization and fine-tuning required to generate novel driving scenarios for training diversity while using 3D Gaussian splatting for the assets allows for both photorealism and high rendering speeds at a relatively low computational cost.

\section{Simulation}

\subsection{Environment}
The environment is a relatively large 3D Gaussian splatting asset where the simulation occurs, and all assets are instantiated within. This approach, compared to the standard mesh-based approach, allows for higher photorealism in the training environment. Novel views and lighting information can be observed by the ego vehicle, all preserved in an asset that is much smaller than the input source or even a classical mesh-based digital twin of the same input scene \cite{Dalal_2024}.

Data intended to be used to capture any asset, including the environment, largely follows best practices for capturing good input data for standard 3D Gaussian splatting \cite{kerbl3Dgaussians}. However, there is a key difference between capturing standard subjects of 3D Gaussian splatting and driving simulator environments. When using single-view RGB images, the input data is no longer a closed-loop sequence. Data captured for 3D Gaussian splatting requires even coverage of the object and sufficient overlapping features between each image. Structure-from-motion (SfM) can match common features between images, resulting in a sufficiently accurate point cloud. Unfortunately, this is not the case when capturing driving scenarios. While sequential pairs of images may have many features in common, the first and final likely do not, provided the input data doesn't include the vehicle driving along the same feature multiple times. More so, the quality of the extraction of common features is usually directly affected by camera frame rate, vehicle speed, lighting conditions, etc.

Using standard data capturing methods with data obtained from a vehicle, we can largely alleviate this issue by using multi-view RGB input \cite{Fei_2024}. When using single-view RGB input, the data does not contain enough views of every feature in the scene. The data also does not contain enough information to place every feature's point clouds accurately in the scene. However, using multi-view data allows us to capture different views of the same features from all sides of the vehicle. This creates common features across multiple images, which hopefully results in enough data for feature extraction and accurate point cloud generation.

Ideal datasets contain multiple views of the environment from the ego vehicle, preferably at regular intervals. At any time $t$ in the simulation, the quality of the scene varies depending on of the input data contained images of the scene at time $t$. If not, then the scene will have to be synthesized directly by the model. As the input data becomes more sparse, this synthesized scene becomes less accurate. So, to preserve quality, a higher frequency of data capture is preferable and yields a better result. The nuScenes dataset \cite{nuscenes} was used in this project for creating the environment.

The nuScenes dataset \cite{nuscenes} provides multi-view images of a vehicle driving through various scenes. Most relevant to our scene reconstruction, the data contains 6 RGB camera streams, all capturing features at the same time. Training the standard 3D Gaussian splatting process with nuScenes data yields results that retain a lot of the features present in the scene but contain a lot of the common issues common with poor quality Gaussian splats, including inaccurate features and "floaters"; artifacts that remain in the image, usually as a result of sparse data.

Once a 3D Gaussian splat is obtained from the data, it must then be converted into a usable Unity asset by using the UnityGaussianSplatting package \cite{Pranckevičius_2023}. This allows for the manipulation of the environment through the Unity 3D engine. The environment's orientation and position can be random due to many factors, including the camera intrinsics and extrinsic, SfM's initialization of the point clouds, how the 3D Gaussian splatting process changes the rotation and position of the point clouds, etc. So, once the asset is imported into Unity, the environment asset is transposed to the origin $(0,0,0)$ of the scene. Using camera extrinsics, the upward vector of the first camera in the scene is used to set the gravity of the engine. That is, the environment's gravity is set to the inverse vector of the camera's Z axis. This allows all assets within the scenes to obey the laws of physics within the environment.


While a 3D Gaussian splat environment includes benefits such as photorealism, relatively simple data capture procedures, fast rendering, and compact storage requirements compared to the traditional mesh-based approach, there are still a lot of improvements that can potentially be made. Quality varies greatly based on how sparse the input set is and the image contains residual artifacts. 3D Gaussian splats also lack relighting capabilities due to the lighting conditions being baked into the scene through the 3D Gaussians' spherical harmonics, unlike a mesh-based environment that reacts to light dynamically in the scene. 

To alleviate some of these issues, a road spline track is created. This is an invisible track that constrains the ego vehicle to areas close to the camera extrinsics. This process ensures that the ego vehicle gets the most accurate view of the scene and provides a way for the ego vehicle to interact with the scene.

\begin{figure}[hbt!]
    \centering
    \begin{subfigure}[t]{0.49\textwidth}
        \centering
        \includegraphics[width=\textwidth, height=120pt]{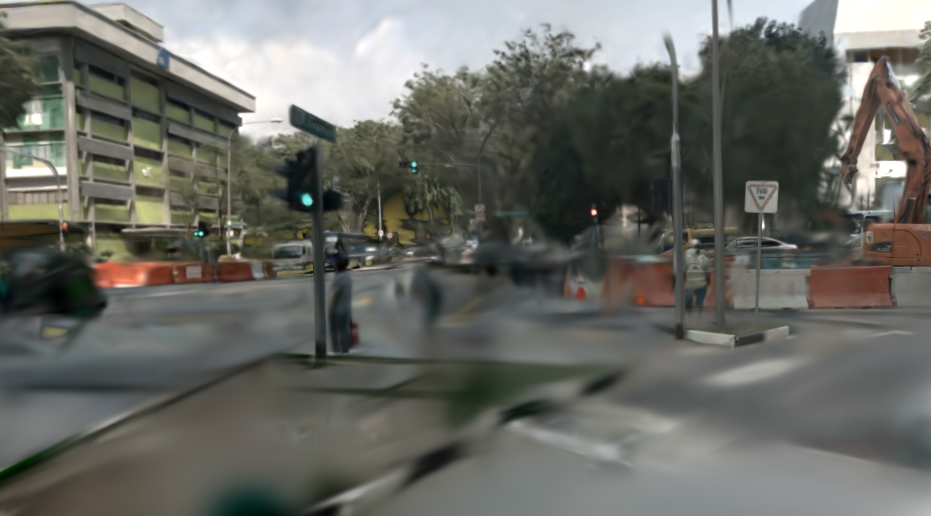}
        \caption{The environment as seen from the ego vehicle while close to the input camera extrinsics.}
    \end{subfigure}
    \hfill
    \begin{subfigure}[t]{0.49\textwidth}
        \centering
        \includegraphics[width=\textwidth, height=120pt]{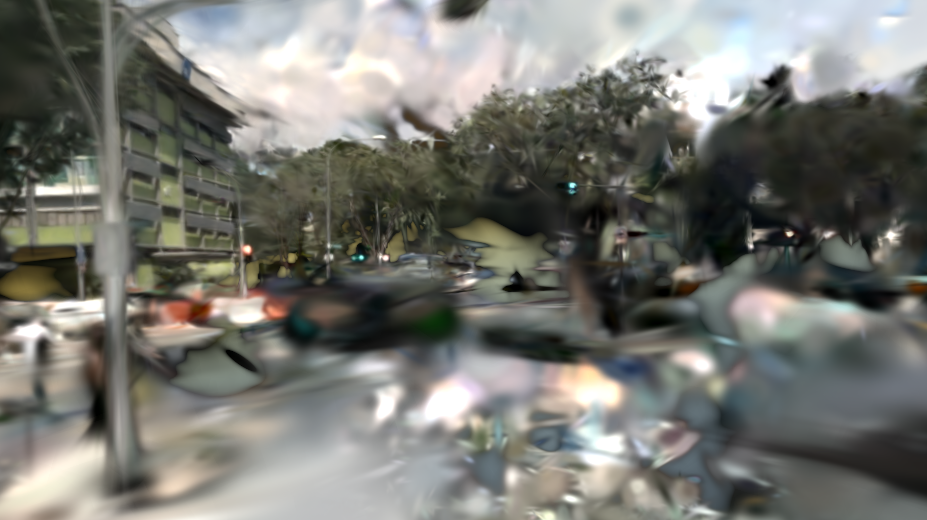}
        \caption{The environment as seen from the ego vehicle while far away from the input camera extrinsics.}
    \end{subfigure}

    \caption{Environment generation from a 3D Gaussian splat of a scene from the nuScenes dataset \cite{nuscenes}}
    \label{fig:road-spline-track-2}
\end{figure}

\subsection{Road Spline Track}
Due to the sparse nature of the input data, the quality of novel views quickly diminishes as the view strays away from the original point of view of the input data. Namely, any scene synthesized from the Gaussian splat of the environment that differs greatly from the point of view of the input data will yield unreliable results. However, since driving simulations are constrained in nature (i.e., both the simulated vehicle and data-capture vehicle need to drive on the road, not drive through buildings, on sidewalks, etc.), the views and, by extension, the positions of the ego vehicle and the data capture vehicle should be roughly similar. That is, the data capture vehicle route constrains where the ego vehicle may go. This observation allows us to utilize the camera extrinsics resulting from the 3D Gaussian splatting process to construct a spline where we know the environment would look the most accurate.


When the simulation is initialized, camera extrinsics from the 3D Gaussian splat asset are processed. A spline is constructed, where each camera position $(X,Y,Z)$ serves as a knot on the spline. This spline effectively serves as the original path that the vehicle took when capturing the input data.

In order to ensure that physics acts on the ego vehicle accurately within the Unity Engine, we need to create a way for the ego vehicle to interact with the environment. Namely, we need to find a way to have the ego vehicle be able to interact with the road, as well as other features of the environment such as buildings, curbs, etc. As 3D Gaussian splats are inaccurate in large unbounded scenes \cite{Jiang2024LIGSGS}, we won't be able to use techniques such as 3D reconstruction to create a physics-accurate scene. Instead, we can use the camera extrinsics provided by the 3D Gaussian splatting process to do this, using the spline created from camera extrinsics as a guide.

A prefabricated asset, called a RoadBlockAsset, is designed as a section of a road with two walls on either side. The width of the road is approximately twice the width of the ego vehicle, and the walls are approximately twice the height of the ego vehicle. This RoadBlockAsset is instantiated repeatedly along the spline, creating a track. The frequency of the RoadBlockAssets is configurable, where a parameter $f$ dictates the spacing between the RoadBlockAssets. A lower $f$ creates a smoother track at the cost of performance. The upward vector of each RoadBlockAsset is equivalent to the upward vector of the camera extrinsics, and the forward vector of rotation is set to the nearest RoadBlockAsset further up the spline. That is, for any RoadBlockAsset $r$, the forward vector directly points to RoadBlockAsset $r+1$. Additionally, all the RoadBlockAssets are offset downward by half the height of the ego vehicle. This is to ensure that the front camera of the ego vehicle lines up perfectly with the spline, which is where the camera extrinsic data places the input cameras. (Figure 3)

To allow the ego vehicle to interact with this environment, a mesh collider is applied to each RoadBlockAsset. The mesh renderer for every block is disabled, effectively creating an invisible track for the ego vehicle to drive within. This track allows the ego vehicle to, from its perspective, appear like it is interacting directly with the 3D Gaussian splat environment. 

\begin{figure}[h]
    \centering
    \begin{subfigure}[t]{0.49\textwidth}
        \centering
        \includegraphics[width=\textwidth, height=120pt]{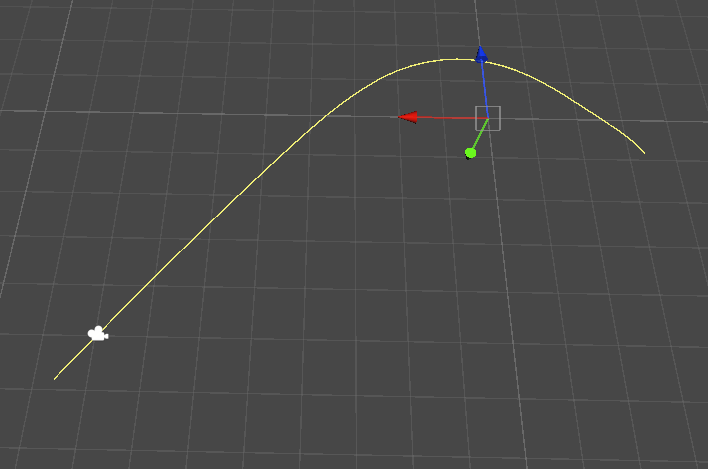}
        \caption{Road spline created from camera extrinsic data. Viewpoints along this spline are closely positioned to the input image views, increasing the chance of accurate views from the ego vehicle.}
    \end{subfigure}
    \hfill
    \begin{subfigure}[t]{0.49\textwidth}
        \centering
        \includegraphics[width=\textwidth, height=120pt]{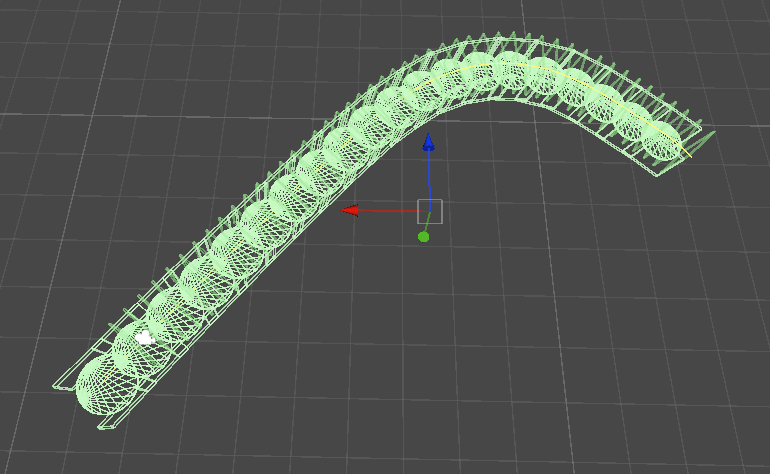}
        \caption{Road track constructed around the road spline, allowing the ego vehicle to physically interact with the scene.}
    \end{subfigure}
        \begin{subfigure}[t]{0.49\textwidth}
        \centering
        \includegraphics[width=\textwidth, height=120pt]{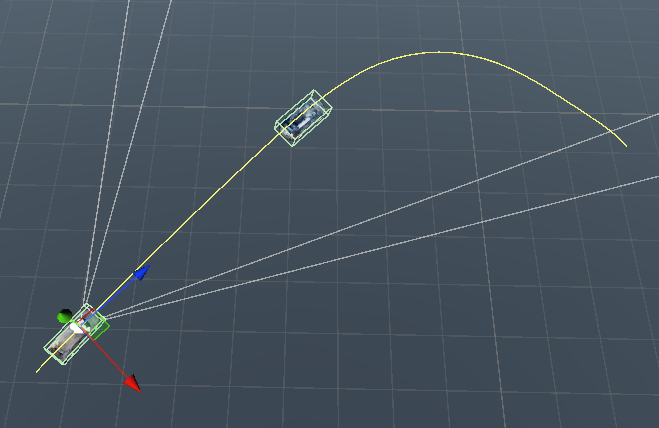}
        \caption{The ego vehicle (bottom left) and a dynamic vehicle agent (top) are instantiated on the spline. Note that the ego vehicle is not attached to the spline in any way but is driving on the track constructed from the spline in Figure 2B.}
    \end{subfigure}
    \hfill
    \begin{subfigure}[t]{0.49\textwidth}
        \centering
        \includegraphics[width=\textwidth, height=120pt]{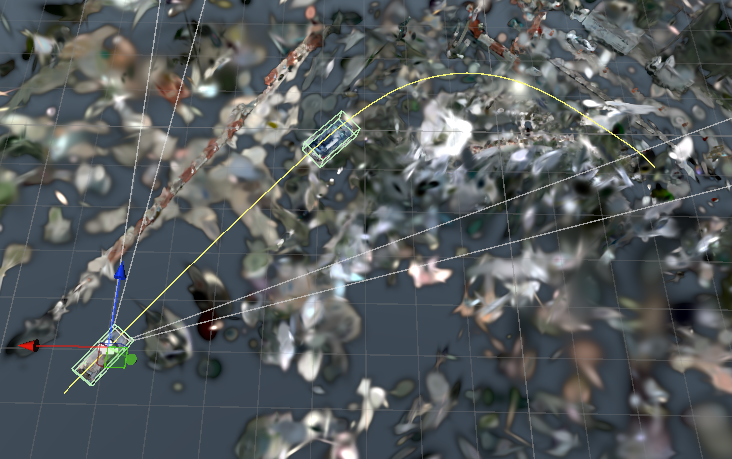}
        \caption{The ego vehicle (bottom) and a dynamic vehicle agent (top) instantiated within the environment asset.}
    \end{subfigure}
    \caption{}
    \label{fig:road-spline-track-1}
\end{figure}

\newpage

\begin{figure}[h]
    \centering
    \includegraphics[width=0.45\textwidth]{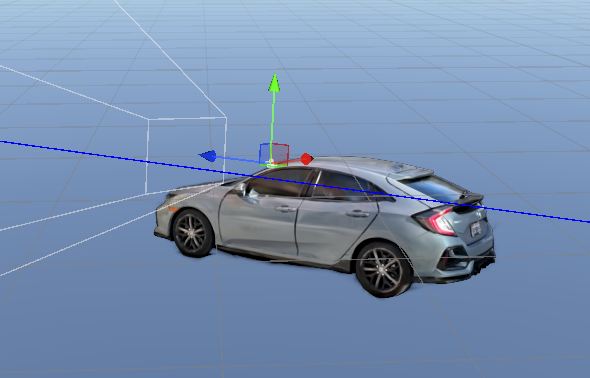}
    \caption{RoadBlockAssets are offset down by half the height of the ego vehicle so that the virtual front camera observes the point of view of the input cameras.}
    \label{fig:spline=car}
\end{figure}

\FloatBarrier
\section{Autonomous Driving} 
\subsection{Ego Agent}
The ego vehicle used in simulations, for all the benefits mentioned above, is also a 3D Gaussian splatting asset. Since capturing the ego vehicle is an easier task for the 3D Gaussian process as it is a bounded scene of an object, it doesn't suffer from the same downfalls as creating an asset of the environment.

3D Gaussian splatting was the chosen method for the other assets for the graphical fidelity, accurate lighting, and compact size compared to a classical texture and mesh asset. However, the asset still captures a large part of the background, so pre-processing is required to clean up the asset. The open-source tool SuperSplat \cite{PlayCanvas_2024} was used to manually clean the assets, removing all 3D Gaussians that don't contribute to the asset's final form. Even with the time taken to clean the assets, it takes considerably less time and resources to create an ego vehicle asset than with classical mesh-based techniques while achieving photorealistic results. More importantly, 3D Gaussian splats take less compute to render at high frame rates.

The ego vehicle asset is functionally different from a standard vehicle model in Unity. Unity renders 3D Gaussian splats as point clouds, and so another level of pre-processing is required to convert the ego vehicle asset into an asset controllable by the Unity engine.

First, a large box collider is attached to the ego vehicle asset. This allows for collision detection between the ego vehicle and the track walls, as well as any collisions with other vehicle agents. Second, four separate wheel colliders are placed in matching locations on the ego vehicle asset. This allows for collision with the RoadTrackAsset so that the ego vehicle remains on the track. The wheel colliders also allow the vehicle to accept inputs, namely, torque vectors and steering angles that move the ego vehicle front, back, left, and right. 

\begin{figure}[hbt!]
    \centering
    \begin{subfigure}[b]{0.4\textwidth}
        \centering
        \includegraphics[width=\textwidth, height=110pt]{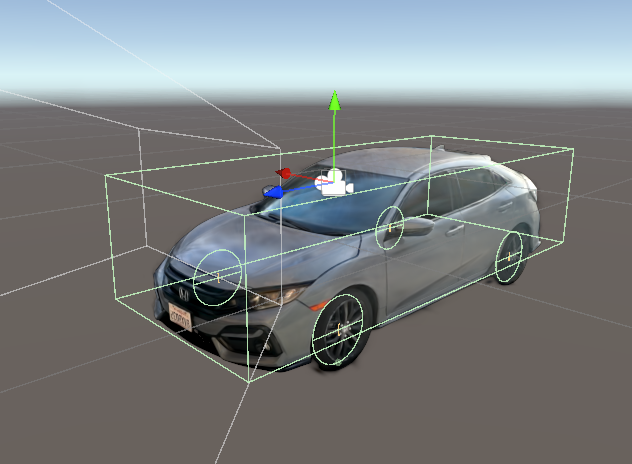}
    \end{subfigure}
    \begin{subfigure}[b]{0.4\textwidth}
        \centering
        \includegraphics[width=\textwidth, height=110pt]{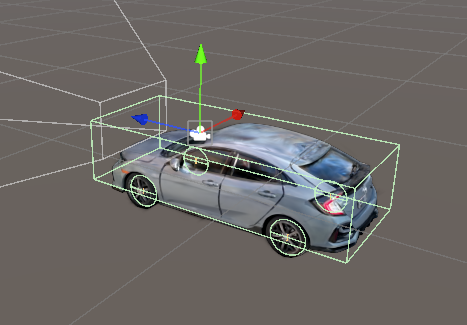}
    \end{subfigure}
    \caption{Example of an ego vehicle asset. Attached to the wheels are the wheel colliders that allow for torque and steering vectors to be applied to the vehicle. A rectangular, configurable collision boundary is attached to the ego vehicle to detect collisions with the environment or other vehicles}
    \label{fig:ego_vehicle-1}
\end{figure}

\subsection{Vehicle Agent}
Similar to the ego vehicle, any agent vehicles in the scene are also instantiated as a 3D Gaussian splat asset. This allows the environment to be seen not only as photorealistic but also as any novel agent vehicle operating in the scene.

The vehicle agent assets are also pre-processed to remove any 3D Gaussians not contributing to the final form of the asset. A tightly bounding box is placed around the vehicle agent so that collisions with the ego vehicle are detected by the model.

A vehicle agent controller determines the dynamics of the vehicle agents in the scene. For any time step $t$, the vehicle agent controller updates the position of the vehicle agents in a forward direction along the spline. The upward vector applied to the agent vehicles is the same upward vector extracted from the camera extrinsics. The forward vector of the vehicle agents aligns with the spline forward vectors at any position $p$ along the spline, allowing agent vehicles to be dynamic within the environment and allowing for the addition of obstacles in the digital twin that weren't originally present in the real world scene.

\begin{figure}[hbt!]
    \centering
    \begin{subfigure}[b]{0.4\textwidth}
        \centering
        \includegraphics[width=\textwidth, height=110pt]{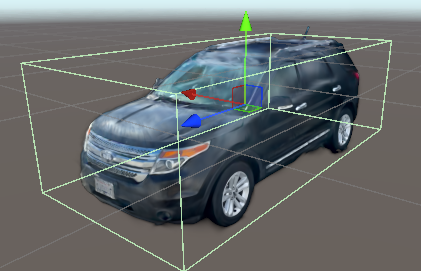}
    \end{subfigure}
    \begin{subfigure}[b]{0.4\textwidth}
        \centering
        \includegraphics[width=\textwidth, height=110pt]{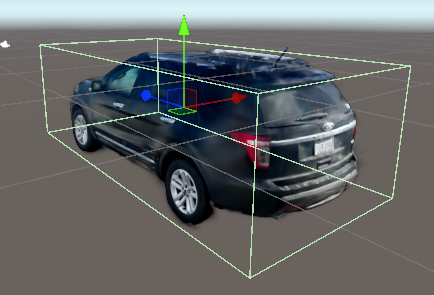}
    \end{subfigure}
    \caption{Example of a vehicle agent asset. The vehicle's position is manipulated for each time-step $t$ along the spline A rectangular, configurable collision boundary is attached to the vehicle agent to detect collisions with the ego vehicle.}
    \label{fig:ego_vehicle-2}
\end{figure}


\section{Experiments}
We evaluate the simulator on 3 different tasks, observing resource usage for each task, as well as model training accuracy. 

\begin{itemize}
  \item No dynamic agents, small scene. The ego vehicle must drive in a straight line from the starting point to the goal.
  \item No dynamic agents, large scene with one turn. The ego vehicle must initially drive in a straight line from the starting point, then turn right and continue driving to the goal. The large scene is roughly twice the drivable area as the small scene.
  \item Dynamic agents, small scene. The ego vehicle must drive in a straight line, but there is one dynamic agent driving in front of the ego vehicle. The ego vehicle must avoid collisions with the dynamic vehicle agent on the way to the goal.
\end{itemize}

The simulator is configured to utilize the ml-agents library\cite{juliani2020}. It is an open-source project that allows for the training of the ego vehicle agent within the environment. The ego vehicle is initialized at one end of the spline, and a goal collider is initialized at the other. An episode of training is designed to end negatively if the ego vehicle collides with either the RoadBlockAsset walls or any agent vehicle. An episode ends positively if the ego agent collides with the goal collider. The ego vehicle uses default parameters for motor force, break force, vehicle mass, steering angle, and drag, as well as any other parameters.

\section{Results}
The training was performed over 250,000 episodes for each task using the standard ml-agents \cite{juliani2020} PPO policy. Table 1 reports the results of letting the ego vehicle train on the three different tasks. We measure the accuracy of the model, seeing how well it performs after training with 50 test episodes. We also measure resource utilization, including the average FPS, average GPU utilization percentage, and average VRAM usage, all as percentages.

While the model performed best with task 1, it still didn't perform perfectly, even with the simple task of driving straight toward the goal. We can see that this isn't an unusual result, as even mesh-based simulators such as CARLA \cite{conf/corl/DosovitskiyRCLK17}, which don't suffer from noisy ego vehicle observations due to photorealism, experience similar results in a similar task.

\begin{table}[hbt!]
    \centering
    \caption{Results of training and testing the model}
    \begin{tabular}{ |c|c|c|c|c| } 
    \hline
        Task & Model Acc. & Avg. FPS & Avg. GPU Util. & Avg. VRAM Usage \\
    \hline
        No dynamic agents, small scene & 86 & 28  & 36.3 & 25.2 \\ 
        No dynamic agents, large scene & 68 & 25  & 41.4 & 31.3 \\ 
        Dynamic agents, small scene    & 81 & 30  & 38.6 & 28.6 \\ 
    \hline
    \end{tabular}
\end{table}

Regarding accuracy, we can see that the model performed best within a small scene, getting above 80 percent accuracy in both cases. We can also observe that when the problem space increased to a larger scene, such as with task 2, the model accuracy dropped to about 70 percent. This can be attributed to a more complex problem space, as the model would need to learn turning parameters in addition to motor and break force parameters.

With resource utilization, due to the simple nature of task 1, we can use its results as a baseline to analyze performance and compare its results to the other two tasks. In task 2, the drivable area for the ego vehicle is roughly double in size, but the increase in average GPU utilization and average VRAM usage has not doubled compared to task 1 and increased by a relatively minor amount. With task 3, despite adding dynamic agents to the scene, we observe that the increase in average GPU utilization and average VRAM usage is marginal, only increasing by at most 3 percent. We also see that average FPS only marginally increases when the drivable area is doubled in size and essentially remains the same when adding dynamic agents to the scene. We can attribute these resource utilization benefits to the efficiency of using 3D Gaussian splatting assets within the simulator. Since 3D Gaussian splatting assets are made of point clouds that take up less space than an identical mesh-based asset, resource utilization reflects this efficiency in asset visualization.

\section{Future Works}
While 3D Gaussian splatting has proven to be a revolutionary new paradigm in novel-view synthesis, many surveys show the drawbacks of the method. 3D Gaussian splatting can produce residual artifacts in the final image, resulting in a less accurate model for an autonomous vehicle training environment \cite{chen2024survey3dgaussiansplatting, Dalal_2024, Fei_2024, bao20243dgaussiansplattingsurvey}. Also, the standard method of 3D Gaussian splatting does not support dynamic Gaussian splats, which means that simulators still need to move dynamic components at runtime. Since most physics engines are not 3D Gaussian splatting native, a level of translation, between a 3D Gaussian splatting asset and a standard asset (in our paper it was converting the 3D Gaussian splat into a Unity asset), still needs to be done. We recognize 4 possible methods to improve this 3D Gaussian splatting simulator, which would vastly improve fidelity, performance, and usability.

Standard 3D Gaussian splatting results in assets that are unable to be relighted with different lighting conditions in 3D engines such as Unity \cite{gao2024relightable3dgaussiansrealistic, Dalal_2024}. However, methods such as Relightable 3D Gaussians 
\cite{gao2024relightable3dgaussiansrealistic} have been developed to result in assets that can achieve photorealistic relighting. This would enable simulators to dynamically adjust the environment to any lighting environment, increasing training diversity and improving the model.

While the environment is a 3D Gaussian splatting asset, any agent that needs to be dynamic to the scene is made as a separate 3D Gaussian splatting asset, initialized into the environment, and dynamically animated within the environment. While this allows for novel driving scenarios to be created within the simulator not initially present in the input data, other dynamic elements are not represented. This can include pedestrians, vehicles on the other side of the road, traffic lights, etc. Dynamic Gaussian splatting methods have been introduced, such as $S^3$Gaussian \cite{huang2024textits3gaussianselfsupervisedstreetgaussians} and DrivingGaussian \cite{zhou2024drivinggaussiancompositegaussiansplatting} that allow for dynamic Gaussians to be present in the scene. The ego vehicle would perceive the environment as a real vehicle would, with multiple dynamic agents in the scene, which may or may not interact with the ego vehicle, further improving accuracy in the model.

Improvements can also be made to the environment to achieve accurate object-level reconstruction. This can be done by utilizing methods such as LI-GS \cite{Jiang2024LIGSGS} and DN-Splatting \cite{turkulainen2024dnsplatter} that allow depth data to be used during training. This can allow the environment to achieve even better alignment with the true scene geometry, drastically reducing errors of reconstructing from sparse input data.

Since constructing the environment proves to be difficult using sparse images, we may utilize novel methods in large-scale scene reconstruction that still utilize 3D Gaussians. SCube \cite{ren2024scube} allows for the reconstruction of large-scale 3D scenes, including the geometry, appearance, and semantics of the scene. This method also allows for text-to-scene generation and LiDAR simulation. Currently, the standard 3D Gaussian splatting model isn't accurate with large unbounded scenes, causing any simulated LiDAR to result in irregular and meaningless output. However, with an accurate 3D scene, the ego vehicle observations can also include LiDAR, allowing for sensor fusion capabilities in the ego vehicle so that our model can gain a better understanding of the scene, resulting in a more accurate model. An accurate 3D reconstruction can also allow us to have dynamic agents within the scene, including the ego vehicle, to interact directly with the scene. We would no longer need the RoadBlockAsset to simulate interactions between assets and the environment, and the decreased overhead of the simulation would result in a performance boost for the model.

\section{Conclusion}
We have presented GSAVS, an autonomous vehicle simulator that allows for the creation and development of autonomous vehicle models where all assets, including the environment, are 3D Gaussian splats. This allows for photo-realism with relatively low compute requirements relative to classical mesh-based simulators. Using a 3D engine, we are able to extract information from the input data and utilize it to create a digital twin of a real-world scene, further augmenting it with 3D Gaussian splat obstacles and increasing training variety to create a more robust model.

\bibliographystyle{plain}
\bibliography{./refs.bib}

\end{document}